%% file: TR_AB_OL_TOF.tex
\documentclass[article]{jss}
\input{Preamble_AB}

\usepackage{array,graphicx,soul,titletoc}
\usepackage{amsfonts,enumitem}
\usepackage[cmex10]{amsmath}
\usepackage{amssymb, cases,balance,cite, lineno,}
\usepackage{booktabs,mathrsfs,bbm,setspace}
\usepackage{rotating}

\DeclareSymbolFont{extraup}{U}{zavm}{m}{n}
\DeclareMathSymbol{\varheart}{\mathalpha}{extraup}{86}
\DeclareMathSymbol{\vardiamond}{\mathalpha}{extraup}{87}

\usepackage{amsthm}

\def\R{\in \mathbb{R}}

\def\T{\top}

\def\j{\jmath}

\def\C{\mathsf{C}}

\def\DE{\stackrel{\mathrm{def}}{=}}

\def\l{\left(}
\def\r{\right)}

\author{Ayush Bhandari\\ \textsf{Massachusetts Institute of Technology} \And Achuta Kadambi \\ \textsf{Massachusetts Institute of Technology} \AND  Refael Whyte \\ \textsf{University of Waikato} and \\ \textsf{Massachusetts Institute of Technology} \And Christopher Barsi \\ \textsf{Massachusetts Institute of Technology} \AND Micha Feigin \\ \textsf{Massachusetts Institute of Technology} \And Adrian Dorrington  \\ \textsf{University of Waikato} \AND Ramesh Raskar \\ \textsf{Massachusetts Institute of Technology}}

\title{\sf Resolving Multi-path Interference in Time-of-Flight Imaging via Modulation Frequency Diversity and Sparse Regularization}

\Plainauthor{Ayush Bhandari} %% comma-separated
\Plaintitle{Resolving multi-path interference in time-of-flight imaging} %% without formatting
\Shorttitle{Resolving Multi-path Interference in Time-of-Flight Imaging} %% a short title (if necessary)

\Abstract{Time-of-flight (ToF) cameras calculate depth maps by reconstructing phase shifts of amplitude-modulated signals. For broad illumination or transparent objects, reflections from multiple scene points can illuminate a given pixel, giving rise to an erroneous depth map. We report here a sparsity regularized solution that separates $K$ interfering components using multiple modulation frequency measurements.  The method maps ToF imaging to the general framework of spectral estimation theory and has applications in improving depth profiles and exploiting multiple scattering.}
 \Volume{39}
 \Issue{06}
 \Month{April}
 \Year{2014}
 \Submitdate{20XX-XX-XX}
 \Acceptdate{20XX-XX-XX}

\Address{
  Ayush Bhandari\\
  Media Laboratory, 75 Amherst St.\\
  Massachusetts Institute of Technology\\
  02139, Cambridge. USA\\
  E-mail: \email{Ayush.Bhandari@googlemail.com}, \email{ayush@MIT.edu} \\
  URL: \url{http://mit.edu/~ayush} \\

  Achuta Kadambi\\
  Media Laboratory, 75 Amherst St.\\
  Massachusetts Institute of Technology\\
  02139, Cambridge. USA\\
  E-mail: \email{achoo@MIT.edu}\\

  Refael Whyte\\
  School of Engineering, University of Waikato\\ 
  Private Bag 3105 \\
  Hamilton 3240. New Zealand  \\
  E-mail: \email{rzw2@students.waikato.ac.nz}\\

  Christopher Barsi\\
  Media Laboratory, 75 Amherst St.\\
  Massachusetts Institute of Technology\\
  02139, Cambridge. USA\\
  E-mail: \email{cbarsi@MIT.edu}\\
  
  Micha Feigin\\
  Media Laboratory, 75 Amherst St.\\
  Massachusetts Institute of Technology\\
  02139, Cambridge. USA\\
  E-mail: \email{michaf@MIT.edu}\\
  
  Adrian Dorrington\\
  School of Engineering, University of Waikato\\ 
  Private Bag 3105 \\
  Hamilton 3240. New Zealand  \\
  E-mail: \email{adrian@waikato.ac.nz}\\

  Ramesh Raskar\\
  Media Laboratory, 75 Amherst St.\\
  Massachusetts Institute of Technology\\
  02139, Cambridge. USA\\
  E-mail: \email{raskar@MIT.edu}
  }

\begin{document}
\begin{figure}[!t]
\centerline{\includegraphics[width=0.5\linewidth]{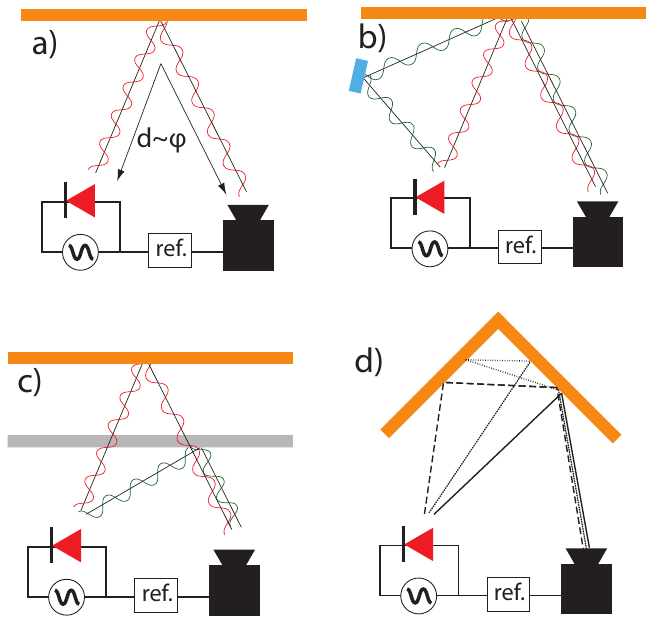}}
\caption{(a)  ToF principle: the phase delay of an emitted AMCW wave proportionally encodes the distance of the reflecting object. (b) Mirror-like and (c) semi-transparent reflections produce MPI at a given camera pixel and yields an incorrect phase.  (c) A complicated scene with severe MPI.}
\label{Fig:1}
\end{figure}

\setlength{\parskip}{15pt}
\newpage
\begin{spacing}{1.2}
Optical ranging and surface profiling have widespread applications in image-guided surgery \cite{cashMP}, gesture recognition  \cite{breuer2007hand}, remote sensing \cite{amannOEng}, shape measurement \cite{theobaltCVPR}, and novel phase imaging \cite{halimehOE}.  Generally, the characteristic wavelength of the probe determines the resolution of the image, making time-of-flight (ToF) methods suitable for macroscopic scenes\cite{foix2011lock,kolb2009time,lee2013time}. Although ToF sensors can be implemented with impulsive sources, commercial ToF cameras rely on the continuous wave approach: the source intensity is modulated at radio frequencies ($\sim$10s of MHz), and the sensor reconstructs the phase shift between the reflected and emitted signals.  Distance is calculated by scaling the phase by the modulation frequency (Fig.~\ref{Fig:1}~(a)).  This method, amplitude modulated continuous wave (AMCW) ToF, offers high SNR in real time.

However, AMCW ToF suffers from multipath interference (MPI) \cite{Bhandari:2013, kadambisig13, godbaz2013understanding, frank2009theoretical, TOFError:2011, Droeschel, dorrington2011separating, godbaz2012closed, godbaz2011understanding}.  Consider, for example, the scenes in Figs.~\ref{Fig:1}~(b,c).  Light rays from multiple reflectors scatter to the observation point.  Each path acquires a different phase shift, and the measurement consists of the sum of these components.  The recovered phase, therefore, will be incorrect. Such ``mixed'' pixels contain depth errors and arise whenever global lighting effects exist.  In some cases (Fig. ~\ref{Fig:1}~(d)), the measurement comprises a continuum of scattering paths.  This can be improved with structured light or mechanical scanning \cite{chenTIP,chenSP}, but these are limited by the source resolution.  Computational optimization \cite{fuchsCVPR,jimCVPR} schemes rely on radiometric assumptions and have limited applicability.

Here, we resolve MPI via sparse regularization of multiple modulation frequency measurements.  The formulation allows us to recast this problem into the general framework of spectral estimation theory \cite{stoica97}.  This contribution generalizes the two-component, dual-frequency approach \cite{godbaz2013understanding, dorrington2011separating, godbaz2012closed}, beyond which the two-component optimization methods fail.  Thus, our method here has two significant benefits.  First, we separate MPI from direct illumination to produce improved depth maps.  Second, we resolve MPI into its components, so that we can characterize and exploit multiple scattering phenomena.  The procedure has two steps: (1) record a scene with multiple modulation frequencies and (2) reconstruct the MPI components using a sparsity constraint.

Consider first the single-component case.  Mathematically, the camera emits the normalized time-modulated intensity $s(t)$\footnote{Here, we consider continuous wave imaging and hence the sinusoidal model, but the discussion is generally applicable to any periodic function.} and detects a signal $r(t)$:
\begin{subequations}
	\begin{align}
 s\left( t \right) = 1+s_{0} \cos \left( {\omega t} \right), t\R  \label{s} \\
 r\left( t \right) = \Gamma(1 + s_{0} \cos \left( {\omega t - \phi } \right)). \label{r}
 	\end{align}
\end{subequations}

\noindent Here, $s_{0}$ and $\Gamma \in [0,1]$ are the signal modulation depth and the reflection amplitude, respectively,  $\omega$ is the modulation frequency, and $\phi$ is the phase delay between the reference waveform $s\l t \r$ and the delayed version $r\l  t\r$. For a co-located source and detector, the distance to the object from the camera is given by the relation $d = c\phi /2\omega$, where $c$ is the speed of light.  

Electronically, each pixel acts as a homodyne detector, measuring the cross-correlation between the reflected signal and the reference.  Denoting the complex conjugate of $f \in \mathbb{C}$ by $f^*$, the cross-correlation of two functions $f$ and $g$ is  
\begin{equation}
 \C_{f,g} \left( \tau \right) \DE \mathop {\lim }\limits_{T  \to \infty } \frac{1}{{2T }}\int_{ - T }^{ + T } {f^*\left( {t + \tau} \right)g\left( t \right)dt}. 
\label{CC}
\end{equation}
Note that infinite limits are approximately valid when the integration window $2T$ is such that $T \gg \omega^{-1}$. A shorter time window produces residual errors, but this is easily avoidable in practice. 
The pixel samples the cross-correlation at discrete times $\tau_q$:

\begin{equation}
\label{m}
     m_\omega\left[q  \right] \DE \C_{s,r}\l \tau_q \r  \stackrel{(\ref{CC})}{=}  \Gamma  \left(1+\frac{{s_{0}^{2} }}{2}\cos ( {\omega \tau_q  + \phi } )  \right). %\notag
\end{equation}

 Using the ``$4$ Bucket Sampling'' technique \cite{foix2011lock}, we calculate the estimated reflection amplitude and the phase, $\widetilde \Gamma ,\widetilde \phi$, using four samples ${\tau _q} = \pi q/2\omega$ with $q =0,...,3$:
\begin{subequations}
	\begin{align}
 &\widetilde \Gamma  = \sqrt {{{\left( {{m_\omega }\left[ 3 \right] - {m_\omega }\left[ 1 \right]} \right)}^2} + \left( {{m_\omega }\left[ 0 \right] - {m_\omega }\left[ 2 \right]} \right)^2} /s_{0}^{2},  \label{G_est} \\
 &\tan \widetilde \phi = \left( {\frac{{{m_\omega }\left[ 3 \right] - {m_\omega }\left[ 1 \right]}}{{{m_\omega }\left[ 0 \right] - {m_\omega }\left[ 2 \right]}}} \right). \label{F_est}
 	\end{align}
\end{subequations}

Therefore, we associate a complex value, $z_\omega$, with a pixel measurement:
\begin{equation}
  z_{\omega} = \widetilde \Gamma e^{\j \widetilde \phi(\omega)}.
  \label{z}
\end{equation} 

Note that these results are formally equivalent to wavefront reconstruction via phase-shifting digital holography \cite{yamaguchi97}.

When multiple reflections contribute to a single measurement, the return signal comprises a sum.  In phasor notation, for $K$ components, 
\begin{equation}
r\left( t \right) =  C_0 + \sum\nolimits_{k = 0}^{K - 1} {{\Gamma_k}{e^{\jmath \left( {\omega t - {\phi _k}\left( \omega  \right)} \right)}}}, 
\end{equation}
where $C_0$ is a constant, ${\phi _k}\left( \omega  \right) = 2d_k\omega /c$, and $\left\{ {{d_k}} \right\}_{k = 0}^{K - 1}$ are $K$ depths at which the corresponding reflection takes place. The reflection amplitude of the $k^\text{th}$ surface is $\Gamma_k$. Each pixel records
\begin{equation}
{m_\omega^K }[q] =  C_0+ \frac{s_{0}^{2}}{2}    e^{\j \omega \tau_q}      \sum\nolimits_{k = 0}^{K - 1} {{\Gamma_k}{e^{\jmath  { {\phi _k}\left( \omega  \right)} }}}.
\end{equation}
Importantly, for a given modulation frequency $\omega_0$ (ignoring a constant DC term), $m_{{\omega _0}}^{K}[\tau_{q}] \propto {\exp{\j {\omega _0}\tau_q }}$, i.e., there is no variation with respect to individual depth components $\left\{ {{\Gamma_k(\omega)},{\phi _k}} \right\}_{k = 0}^{K - 1}$ \cite{Bhandari:2013}, regardless of the sampling density.  Equivalently, the camera measurement,

\begin{equation}
  z_{\omega}^{(K)} = \widetilde \Gamma(\omega) e^{\j \widetilde \phi(\omega)} = \sum\nolimits_{k = 0}^{K - 1} {{\Gamma_k(\omega)}{e^{\j { {\phi _k}\left( \omega  \right)} }}}
\end{equation}
is now a complex sum of $K$ reflections, which cannot be separated without independent measurements.  Thus, at a given frequency, the measured phase, and hence the depth, is a nonlinear mixture of all interefering components.

Our method separates these components by recording the scene with equi-spaced frequencies $\omega = n\omega_0$ ($n  \in \mathbb{N}$) and acquiring a set of measurements $\mathbf{z}$: 
\begin{equation}
{\mathbf{z}} = {\left[ {z_{\omega_0}^{(K)},z_{{2\omega _0}}^{(K)}, \ldots ,z_{N{\omega _0}}^{(K)}} \right]^\T}.
\end{equation}
The forward model can be written compactly in vector-matrix form as ${\mathbf{z}} = \boldsymbol{\Phi g} +\boldsymbol{\sigma}$, where $\boldsymbol{\Phi} \in \mathbb{C}^{N\times K}$  is identified as a Vandermonde matrix, 

\begin{equation}
\boldsymbol{\Phi} =
 \begin{pmatrix}
  e^{\j {\omega _0}{\phi _0}} & e^{\j {\omega _0}{\phi _1}} & \cdots & e^{\j {\omega _0}{\phi _{K-1}}} \\
  e^{\j 2{\omega _0}{\phi _0}} & e^{\j2 {\omega _0}{\phi _1}} & \cdots & e^{\j 2{\omega _0}{\phi _{K-1}}} \\
  \vdots  & \vdots  & \ddots & \vdots  \\
    e^{\j N{\omega _0}{\phi _0}} & e^{\j N {\omega _0}{\phi _1}} & \cdots & e^{\j N{\omega _0}{\phi _{K-1}}} \\
 \end{pmatrix},
\label{vandermonde}
\end{equation}

\noindent $\boldsymbol g   = {\left[ {{\Gamma_0}, \ldots ,{\Gamma _{K - 1}}} \right]^\T}\in \mathbb{R}^{K\times 1}$, and $\boldsymbol{\sigma}$ represents zero-mean Gaussian i.i.d. noise, which controls the error $\varepsilon_0$ in our reconstruction algorithm. Our goal is to estimate the phases $\boldsymbol\phi  = {\left[ {{\phi _0}, \ldots ,{\phi _{K - 1}}} \right]^\T}\in \mathbb{R}^{K\times 1}$ and the reflection amplitude vector $\boldsymbol{g}$. 

To recover these quantities, first note the similarity between $\boldsymbol\Phi$ and an oversampled $N \times L$ discrete Fourier transform (DFT) matrix $\boldsymbol\Psi$, with elements $\Psi_{nl} = \exp(\j nl/L)$.  If $L \gg K$, the discretization of $\boldsymbol{\Psi}$ is small enough to assume that the columns of $\boldsymbol\Phi$ are contained in $\boldsymbol\Psi$.  We can also define a vector $\boldsymbol g' \in \mathbb{R}^{L\times 1}$, whose elements are zero except for $K$ reflection amplitudes $\left\{ {{\Gamma_k}} \right\}_{k = 0}^{K - 1}$, such that $\mathbf{z} = \boldsymbol\Psi \boldsymbol g'$. We use the ($K$-)sparsity of $\boldsymbol g'$ to regularize the problem:

\begin{equation}
 \qquad \underbrace {\left\| {{\mathbf{z}} - \boldsymbol{\Psi g'}} \right\|_{{\ell _2}}^2}_{{\textsf{Data-Fidelity}}} < {\varepsilon _0}  \quad \textrm{ such that} \quad \underbrace {{{\left\| {\boldsymbol g'}  \right\|}_{{\ell _0}}} = K}_{{\textsf{Sparsity}}},
\label{OMP}
 \end{equation}

where the $\ell_p$--norm as $\left\| {\mathbf{x}} \right\|_{{\ell _p}}^p \DE \sum\nolimits_n {{{\left| {{x_n}} \right|}^p}}$. The case of $p\rightarrow 0$ is used to define ${{\left\| {\boldsymbol g'}  \right\|}_{{\ell _0}}}$ as the number of nonzero elements of $\boldsymbol g'$. Eq.~\ref{OMP} demands a least-squares solution to the data-fidelity problem $\left\| {{\mathbf{z}} - \boldsymbol{\Psi g'} } \right\|_{{\ell _2}}^2$ up to some error tolerance $\varepsilon _0$, with the constraint that we accommodate up to $K$ nonzero values of $\boldsymbol g'$. 

The sparsity of $\boldsymbol g'$ arises from two underlying assumptions.  First, we do not consider the case of volumetric scattering, which would preclude discrete reflections and require a different parametrization (e.g., through the diffusion coefficient).  Second, we ignore the contributions of inter-reflections between scattering layers, as their amplitudes fall off quickly.  They could be incorporated, into our formulation, with the result of changing the sparsity of $\boldsymbol g'$ from $K$ to $K'$, where $K'-K$ is the number of inter-reflections considered.

We solve Eq.~\ref{OMP} via orthogonal matching pursuit (OMP), which is an iterative algorithm that searches for the best-fit projections (in the least-squares sense) of the coefficients onto an over-complete dictionary.  We input $\boldsymbol \Psi$ and measurements $\mathbf{z}$ into the algorithm.  The outputs are the set of reflection coefficients $\Gamma_k$ and their positions in $\boldsymbol {g'}$. With the position of each $\Gamma_k$ reconstructed, the corresponding phases $\phi_k$ are recovered through the elements of $\boldsymbol{\Psi}$: $\phi_k = (\j n)^{-1}\log(\Psi_{n l_{k}}) = l_{k}/L$, where $l_{k}$ is the location of $\Gamma_k$ in $\boldsymbol g'$.

\begin{figure}[!t]
\centerline{\includegraphics[width=0.8\linewidth]{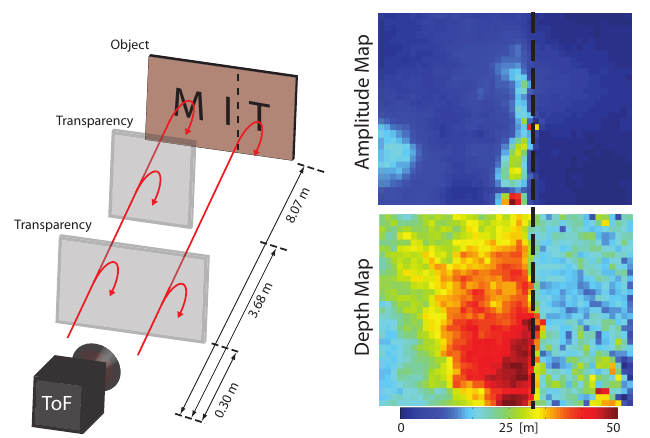}}
\caption{Left: experimental setup.  Two transparencies block the left side of the camera (for a three-component measurement), and one transparency blocks the right (two-component measurement).  Right: measured amplitude and depth at $\omega = 3\omega_0$. Dashed line indicates edge of second transparency.}
\label{Fig:2}
\end{figure}

We verify this theory with the experimental setup shown in Fig.~\ref{Fig:2}.  A PMD19k-2 $160 \times 120$ sensor array is controlled by a Stratix III FPGA.  Analog pixel values are converted to 16-bit unsigned values by an ADC during the pixel readout process.  Eight 100 mW Sony SLD 1239JL-54 laser diodes illuminate the scene.  The lasers are placed symmetrically around the detector for a coaxial configuration.  The base frequency modulation is $f_0 = \omega_{0}/(2\pi) = 0.7937 \textrm{ MHz}$, and the integration time is 47 ms.  The scene consists of three layers.  Farthest, at 8.1 m, is an opaque wall with gray-scale text (``MIT'') printed on it.  Closest, at 0.3 m is a semi-transparent sheet.  Between the two layers is another semi-transparent sheet that covers only the left half of the field of view.  Therefore, the left-hand side records three bounces and the right only two. All three layers are within the depth of field of the camera to avoid mixed pixels from blurring.  

Depth and amplitude maps acquired at a specific frequency are shown in Fig. ~\ref{Fig:2}.  Due to MPI, the measured depths do not correspond to any physical layer in the scene.  All depth and amplitude information from the three scene layers is mixed nonlinearly into a set of composite measurements (pixels) and cannot be recovered.

\begin{figure}[!t]
\centerline{\includegraphics[width=0.6\linewidth]{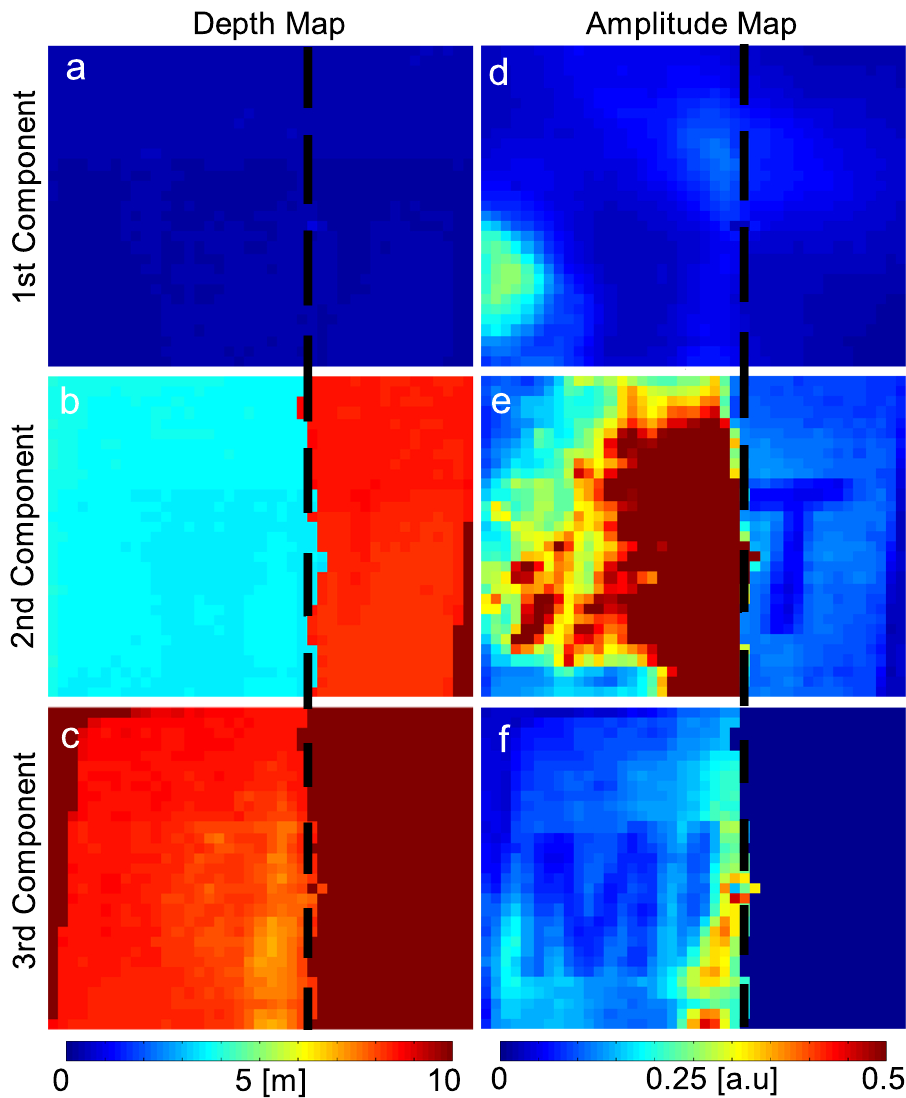}}
\caption{Reconstructed amplitudes and depths via sparse regularization. Dashed lines indicate edge of second transparency.}
\label{Fig:3}
\end{figure}
We repeat the acquisition $77$ times, with modulation frequencies spaced $0.7937 \textrm{ MHz}$ apart and input these data into the OMP algorithm with $K = 3$.  The reconstruction, shown in Fig.~\ref{Fig:3}, shows each depth correctly recovered.  The closest depth map (Fig.~\ref{Fig:3}~(a), first transparency) is constant. The second map (Fig.~\ref{Fig:3}~(b)) contains two depths: the second transparency on the LHS and the wall on the RHS. The third depth map contains the wall depth on the LHS (Fig.~\ref{Fig:3}~(c)).  The third-bounce amplitude (Fig.~\ref{Fig:3}~(f)) is zero where there are only two layers (RHS).  The depth here is therefore undefined, though we set the distance to be 10 m to avoid random fluctuations.  Further, the text is recovered properly in the amplitude maps corresponding to the correct depths (Figs.~\ref{Fig:3}~(e,f)). Note that accurate depths are recovered even in the presence of strong specularity (Fig.~\ref{Fig:3}~(e)).

\begin{figure}[!t]
\centerline{\includegraphics[width=0.8\linewidth]{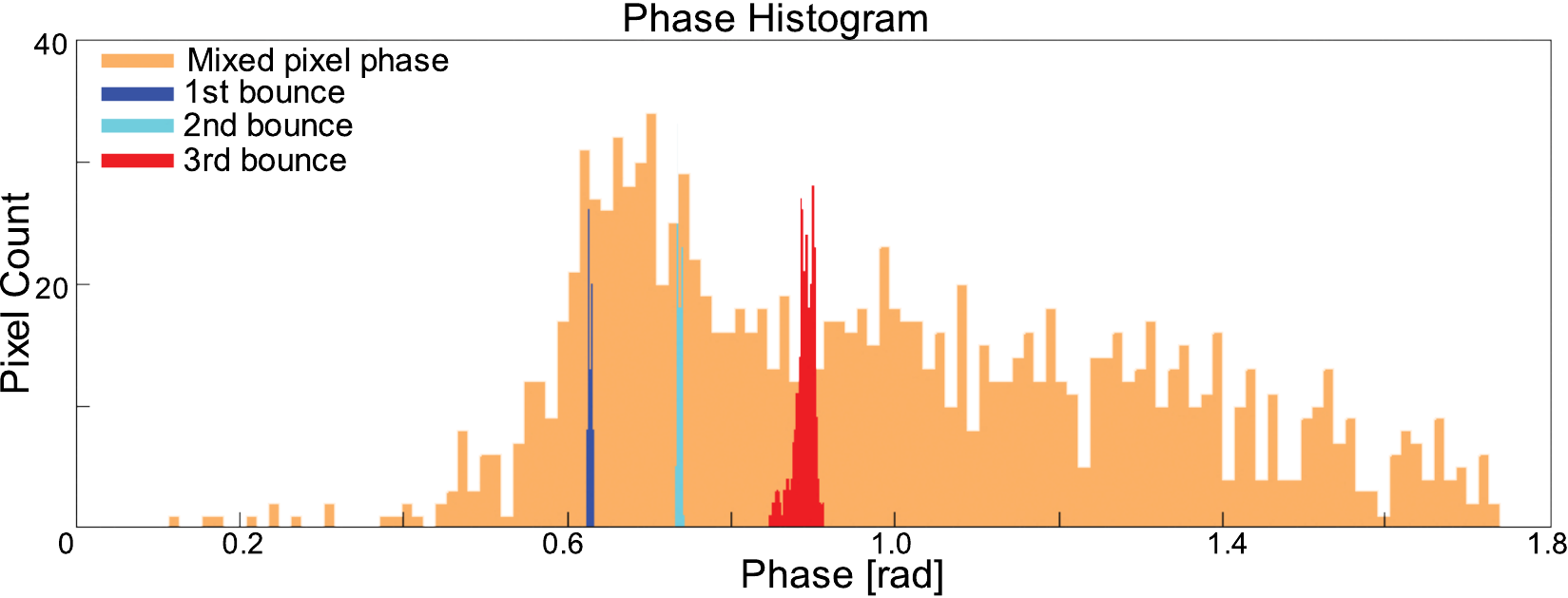}}
\caption{Phase histogram for reconstructed and measured depth maps. Reconstructed phases cluster around the correct depths, whereas the measured depth map has a wide variance across the entire range.}
\label{Fig:4}
\end{figure}

A phase histogram is shown in Fig.~\ref{Fig:4}.  The histogram from the single frequency measurement in Fig.~\ref{Fig:1} varies from 0.6 to 1.8 rad.  Recovered phases are centered around the ground truth values.  The third-phase variance is wider because OMP computes the first two components, leaving little residual energy, so that several columns in $\boldsymbol{\Psi}$ can minimize the least-squares error.

In principle, the technique can be extended to any number of bounces, provided enough modulation frequencies are used (though a first-principles derivation is beyond the scope of this contribution).  In practice, however, the reflected amplitudes decrease with increasing component number, so that higher-order components diminish in importance.  Furthermore, OMP need not assume a number of components that is the same as that of the physical implementation.  If the assumed number is greater than the physical number, OMP will reconstruct all the physical components, with higher-order ones having an amplitude on order of the system noise.  Conversely, if the assumed number is less than the physical number, OMP will recover the strongest reflections.

Therefore, the method is a generalization of global/direct illumination separation and can decompose different elements of global lighting.  This is useful not only for improved depth accuracy, but also imaging in the presence of multiple scatterers such as diffuse layers, sediment, turbulence, and turbid media, as well as in places where third-component scattering must be extracted \cite{veltenNC}.  Furthermore, because it is based on phase measurements, this technique can be mapped to multiple scattering in holography \cite{bartonPRL} by substituting optical frequency for the modulation frequency.

In conclusion, we implemented a multi-frequency approach for decomposing multiple depths for a ToF camera.  The result is general and holds for any number of bounces, and it can be extended to non-harmonic signals \cite{kadambisig13}.  Future work includes calculating bounds on measurements and resolution.  The method can be incorporated with structured illumination and pixel correlations and for edge detection, and refocusing. The result holds promise for mitigating and exploiting multipath for a wide variety of scenes.

%In this section, please provide full versions of citations to assist reviewers and editors (OL publishes a short form of citations) or any other information that would aid the peer-review process.

\footnotesize
\bibliographystyle{plain}
\bibliography{TOF_OL_refs}
\end{spacing}
\footnotesize
\setlength{\parskip}{5pt}
\newpage
\end{document}

%% file: Preamble_AB.tex
\usepackage{array,graphicx}
\usepackage{amsfonts,cite}
\usepackage[cmex10]{amsmath}
\usepackage{cases,balance,cite, lineno}
\usepackage{booktabs,mathrsfs,bbm,amssymb}
\usepackage[table]{xcolor}
\DeclareSymbolFontAlphabet{\mathbb}{AMSb}

\usepackage{setspace}
\definecolor{B1}{HTML}{6666FF}
\definecolor{Red1}{HTML}{FF0033}
\definecolor{LinkRed1}{HTML}{CC0066}
\newcommand{\noprint}[1]{}

\usepackage{amsthm}

\input cyracc.def

\def\DE{\stackrel{\mathrm{def}}{=}}

\def\R{\in \mathbb{R}}

\def\T{\top}
\def\j{\jmath}

\def\C{\mathsf{C}}

\def\l{\left(}
\def\r{\right)}

%% file: TR_AB_OL_TOF.bbl
\begin{thebibliography}{10}

\bibitem{amannOEng}
M.~C. Amann, T.~Boch, R.~Myllyla, M.~Rioux, and M.~Lescure.
\newblock Laser ranging: a critical review of usual techniques for distance
  measurement.
\newblock {\em Opt. Eng.}, 40:10--19, 2001.

\bibitem{bartonPRL}
J.~J. Barton.
\newblock Removing multiple scattering and twin images from holographic images.
\newblock {\em Phys. Rev. Lett.}, 67:3106--3109, 1991.

\bibitem{Bhandari:2013}
Ayush Bhandari, Achuta Kadambi, Refael Whyte, Lee Streeter, Christopher Barsi,
  Adrian Dorrington, and Ramesh Raskar.
\newblock Multifrequency time of flight in the context of transient renderings.
\newblock In {\em ACM SIGGRAPH 2013 Posters}, number~46, 2013.

\bibitem{breuer2007hand}
Pia Breuer, Christian Eckes, and Stefan M{\"u}ller.
\newblock Hand gesture recognition with a novel {IR} time-of-flight range
  camera--a pilot study.
\newblock In {\em Computer Vision/Computer Graphics Collaboration Techniques},
  pages 247--260. Springer, 2007.

\bibitem{cashMP}
D.~M. Cash, T.~K. Sinha, W.~C. Chapman, H.~Terawaki, B.~M. Dawant, R.~L.
  Galloway, and M.~I. Miga.
\newblock Incorporation of a laser range scanner into image-guided liver
  surgery: surface acquisition, reigsration, and tracking.
\newblock {\em Med. Phys.}, 30:1671--1682, 2003.

\bibitem{chenTIP}
S.~Y. Chen, Y.~F. Li, and J.~W. Zhang.
\newblock Vision processing for realtime 3d data acquisition based on coded
  structured light.
\newblock {\em IEEE Trans. Image Proc.}, 17:167--176, 2008.

\bibitem{theobaltCVPR}
Y.~Cui, S.~Schoun, D.~Chan, S.~Thrun, and C.~Theobalt.
\newblock 3d shape scanning with a time-of-flight camera.
\newblock In {\em Proc. Computer Vision and Pattern Recognition}, 2010.

\bibitem{dorrington2011separating}
A.A. Dorrington, J.P. Godbaz, M.J. Cree, A.D. Payne, and L.V. Streeter.
\newblock Separating true range measurements from multi-path and scattering
  interference in commercial range cameras.
\newblock In {\em IS\&T/SPIE Electronic Imaging}, pages 786404--786404, 2011.

\bibitem{Droeschel}
D.~Droeschel, D.~Holz, and S.~Behnke.
\newblock Multi-frequency phase unwrapping for time-of-flight cameras.
\newblock In {\em IEEE/RSJ International Conference on Intelligent Robots and
  Systems (IROS)}, pages 1463--1469, 2010.

\bibitem{foix2011lock}
Sergi Foix, Guillem Alenya, and Carme Torras.
\newblock Lock-in time-of-flight (tof) cameras: a survey.
\newblock {\em IEEE Sensors Journal}, 11(9):1917--1926, 2011.

\bibitem{chenSP}
Active Sensor~Plnanning for Multiview Vision~Tasks.
\newblock {\em Chen, S. Y.}
\newblock Springer, 2008.

\bibitem{frank2009theoretical}
Mario Frank, Matthias Plaue, Holger Rapp, Ullrich K{\"o}the, Bernd J{\"a}hne,
  and Fred~A Hamprecht.
\newblock Theoretical and experimental error analysis of continuous-wave
  {T}ime-of-{F}light range cameras.
\newblock {\em Proc. SPIE Conf. on Vis. Commun. and Image Proc.},
  48(1):013602---013602, 2009.

\bibitem{fuchsCVPR}
S.~Fuchs.
\newblock Multipath interference compensation in time-of-flight camera images.
\newblock In {\em Proc. Computer Vision and Pattern Recognition}, 2010.

\bibitem{godbaz2011understanding}
John~P Godbaz, Michael~J Cree, and Adrian~A Dorrington.
\newblock Understanding and ameliorating non-linear phase and amplitude
  responses in amcw lidar.
\newblock {\em Remote Sensing}, 4(1):21--42, 2011.

\bibitem{godbaz2013understanding}
John~P Godbaz, Adrian~A Dorrington, and Michael~J Cree.
\newblock Understanding and ameliorating mixed pixels and multipath
  interference in amcw lidar.
\newblock In {\em TOF Range-Imaging Cameras}, pages 91--116. Springer, 2013.

\bibitem{godbaz2012closed}
J.P. Godbaz, M.J. Cree, and A.A. Dorrington.
\newblock Closed-form inverses for the mixed pixel/multipath interference
  problem in amcw lidar.
\newblock In {\em IS\&T/SPIE Electronic Imaging}, pages 829618--829618, 2012.

\bibitem{halimehOE}
J.~C. Halimeh and M.~Wegener.
\newblock Time-of-flight imaging of invisibility cloaks.
\newblock {\em Opt. Express}, 20:63--74, 2012.

\bibitem{lee2013time}
Miles Hansard, Seungkyu Lee, Ouk Choi, and Radu Horaud.
\newblock {\em Time-of-flight cameras: principles, methods and applications}.
\newblock Springer, 2013.

\bibitem{jimCVPR}
D.~Jimenez, D.~Pizarro, M.~Mazo, and S.~Palazuelos.
\newblock Modelling and correction of multipath interference in time of flight
  cameras.
\newblock In {\em Proc. Computer Vision and Pattern Recognition}, 2012.

\bibitem{TOFError:2011}
A.~P~P Jongenelen, D.~G. Bailey, A.~D. Payne, A.~A. Dorrington, and D.~A.
  Carnegie.
\newblock Analysis of errors in tof range imaging with dual-frequency
  modulation.
\newblock {\em IEEE Trans. on Instrumentation and Measurement},
  60(5):1861--1868, 2011.

\bibitem{kadambisig13}
A.~Kadambi, R.~Whyte, A.~Bhandari, L.~Streeter, C.~Barsi, A.~A. Dorrington, and
  R.~Raskar.
\newblock Coded time of flight cameras: sparse deconvolution to address
  multipath interference and recover time profiles.
\newblock {\em ACM Trans. Graph.}, to appear.

\bibitem{kolb2009time}
Andreas Kolb, Erhardt Barth, Reinhard Koch, and Rasmus Larsen.
\newblock {T}ime-of-{F}light sensors in computer graphics.
\newblock In {\em Proc. Eurographics}, pages 119--134, 2009.

\bibitem{stoica97}
P.~Stoica and R.~L. Moses.
\newblock {\em Introduction to Spectral Analysis}.
\newblock Prentice Hall, 1997.

\bibitem{veltenNC}
Andreas Velten, Thomas Willwacher, Otkrist Gupta, Ashok Veeraraghavan, M.~G.
  Bawendi, and Ramesh Raskar.
\newblock Recovering three-dimensional shape around a corner using ultrafast
  time-of-flight imaging.
\newblock {\em Nat. Commun.}, 3:745, 2012.

\bibitem{yamaguchi97}
T.~Yamaguchi, I.and~Zhang.
\newblock Phase shifting digital holography.
\newblock {\em Opt. Lett.}, 22:1268, 1997.

\end{thebibliography}
